\documentclass[10pt,twocolumn,letterpaper]{article}
\usepackage[pagenumbers]{cvpr}
\usepackage{graphicx}
\usepackage{float}
\definecolor{cvprblue}{rgb}{0.21,0.49,0.74}
\usepackage[breaklinks,colorlinks,allcolors=cvprblue]{hyperref}
\title{APS-VLA: Action-Guided and Future-Aware Token Pruning for Efficient Vision-Language-Action Models}
\hypersetup{pdftitle={APS-VLA: Action-Guided and Future-Aware Token Pruning for Efficient Vision-Language-Action Models},pdfauthor={Xiaoyang Xing, Anthony Cohn, Chun-Mei Feng, Xiaoliang Gong, Asoke Nandi}}
\author{%
\normalsize Xiaoyang Xing$^1$\quad Anthony Cohn$^2$\quad Chun-Mei Feng$^{1,4,*}$\quad Xiaoliang Gong$^1$\quad Asoke Nandi$^3$\\[3pt]
\small $^1$School of Computer Science and Technology, Tongji University\\
\small $^2$School of Computer Science, University of Leeds\quad
$^3$Electronic and Electrical Engineering, Brunel University of London\\
\small $^4$Agency for Science, Technology and Research (A*STAR), Singapore\\[3pt]
{\footnotesize\texttt{augustusper@tongji.edu.cn}, \texttt{A.G.Cohn@leeds.ac.uk},
\texttt{gxllshsh@tongji.edu.cn}, \texttt{Asoke.Nandi@brunel.ac.uk}}\\[2pt]
{\footnotesize $^*$Corresponding author: \texttt{chunmeifeng@tongji.edu.cn}.}}
\begin{document}
\maketitle

\begin{abstract}
Vision-language-action (VLA) policies have demonstrated strong generalization in robotic manipulation, yet processing dense visual tokens often incurs high inference latency. Visual token pruning offers a practical way to reduce this cost, but identifying a compact set of tokens that preserves reliable action generation remains challenging. Existing methods largely rely on static or immediate heuristic cues to estimate token importance. We introduce APS-VLA, a framework that learns visual token selection from robot demonstrations while keeping the pretrained VLA backbone frozen. A lightweight Decision Influence Estimator scores tokens conditioned on current visual observations, language instructions, and robot proprioception. Its training combines two complementary signals: action supervision matches both the full-input policy predictions and demonstrated actions, while an auxiliary visual transition objective uses demonstrated actions to predict subsequent visual features from the selected representation. At deployment, the estimator performs a single hard selection and packing step before the policy's native action-generation procedure; the auxiliary predictor is removed. Experiments with OpenVLA-OFT on LIBERO and with $\pi_{0.5}$ and XR-1 on RoboCasa365 demonstrate inference acceleration with modest changes in task success. Across five real-world tasks on two dual-arm platforms, APS-VLA reduces mean policy inference latency from 137.25 to 93.25~ms, achieving a $1.47\times$ speedup with an average success rate of 61\%, compared with 60\% for the unpruned baseline. Together, these results support APS-VLA as a modular, plug-in framework for efficient robotic control across VLA architectures, preserving the underlying policy parameters and native action-generation procedure.
\end{abstract}

\begin{figure}[!t]
  \centering
  \makebox[\linewidth][r]{\includegraphics[width=246.89bp,height=166.49bp]{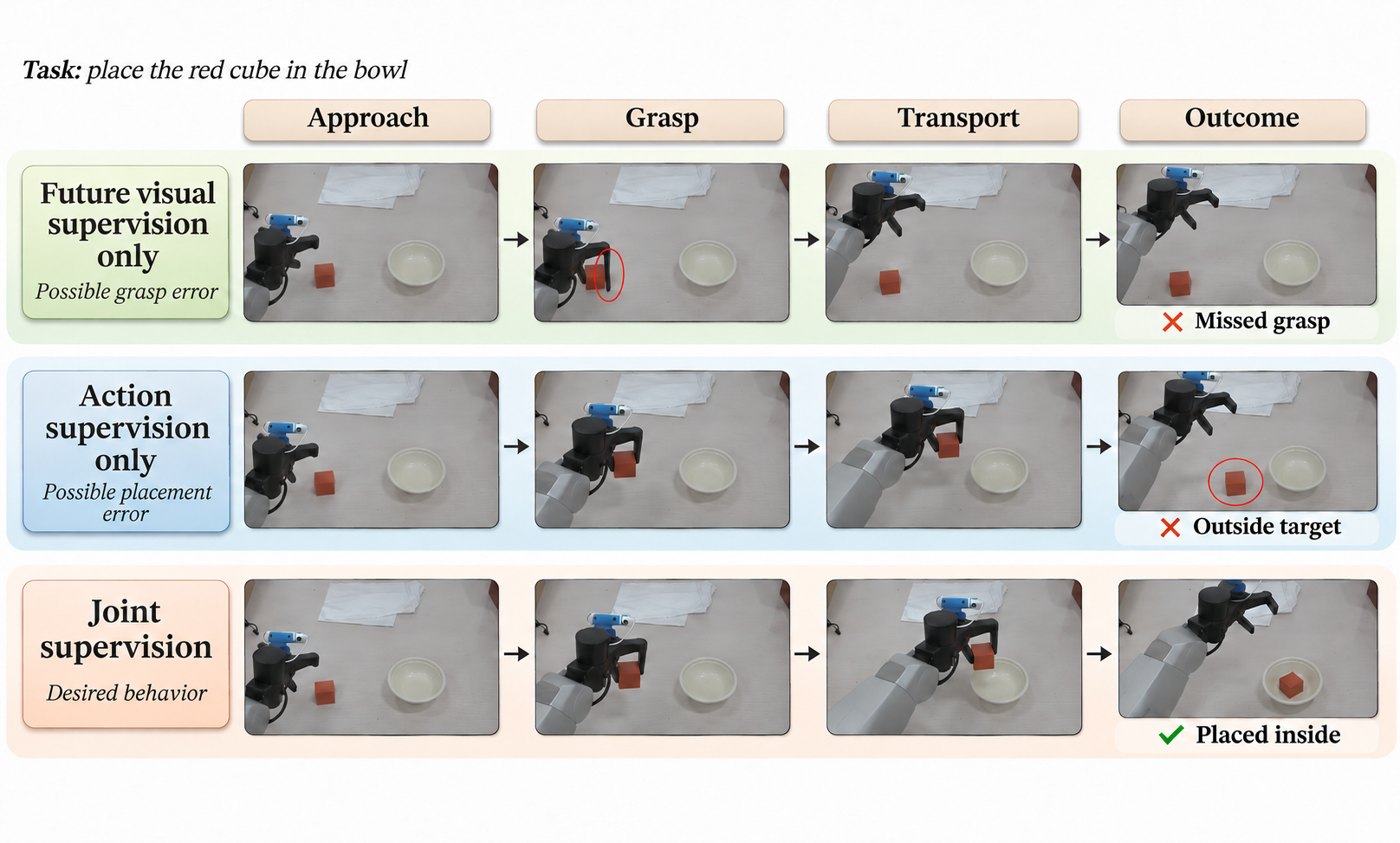}}
  \caption{\textbf{Why visual token selection needs both action and future visual cues.} Placing a cube in a bowl requires both precise gripper--cube alignment and accurate target release. Future visual supervision alone (top) misses fine-grained contact cues and fails to grasp, whereas action supervision alone (middle) secures the grasp but loses destination context, releasing outside the bowl. Joint supervision (bottom) retains both cues for successful \mbox{placement}.}
  \label{fig:motivation}
\end{figure}

\section{Introduction}
\label{sec:introduction}

Vision-language-action (VLA) policies use camera observations, language instructions, and robot state to generate actions across a wide range of manipulation tasks. Models such as OpenVLA, OpenVLA-OFT, and $\pi_{0.5}$ achieve this through different action-generation architectures, yet all must repeatedly process visual observations as the robot moves~\cite{kim2024openvla,kim2025oft,intelligence2025pi05}. With multiple cameras, a policy may process hundreds of visual tokens at each decision step, even when the next action depends on a small part of the scene. Reducing this sequence can lower inference latency. The fundamental challenge lies not merely in identifying redundant tokens, but in how to capture the dynamically evolving visual cues essential for control as a task progresses, while keeping deployment efficient and non-invasive to the pretrained policy.

Figure~\ref{fig:motivation} illustrates why both signals matter when placing a cube into a bowl. Action supervision encourages retaining cues needed to generate the commanded motion, such as the gripper--object relationship during grasping. Future visual supervision complements this constraint by encouraging the selected representation to preserve information about the resulting scene change, such as the object's placement relative to the bowl. Together, they guide token selection toward both immediate control and the visual consequences of actions.

Existing pruning methods largely rely on static spatial redundancy, immediate instruction relevance, or heuristic importance measures~\cite{chen2024fastv,yang2025visionzip}. Although some methods incorporate temporal consistency or adapt pruning to the motion and stage of execution~\cite{xu2025vlacache,liu2025vlapruner,pei2026vlaadp,wang2025specprune}, these criteria do not necessarily preserve the task-relevant visual information needed as manipulation unfolds. Approaches that fine-tune the entire VLA backbone additionally incur substantial training and memory costs. These limitations motivate learning a lightweight selector from robot demonstrations that identifies control-relevant visual cues across task stages under a fixed token budget, while keeping the pretrained policy frozen.

We introduce APS-VLA, a framework that learns visual token selection from paired actions and observations while keeping the VLA backbone frozen. Two training objectives act on the same selected representation. The first requires the frozen policy to reproduce its predictions under full visual input and to match the demonstrated actions. The second uses the selected current representation and the demonstrated action sequence to predict a feature summary of a later observation.

At the center of APS-VLA is the Decision Influence Estimator (DIE), our lightweight visual token scorer (Fig.~\ref{fig:die}). DIE combines the features of each image patch with visual context, the instruction, and normalized robot proprioception. Both objectives train this scorer through differentiable selection. At deployment, DIE scores the current tokens once, retains the highest-scoring tokens within each camera view, and packs them for the frozen policy. The future observation and demonstrated actions are used only during training. The deployed policy uses its original action head and requires no online prediction of future observations.

\begin{figure*}[t]
  \centering
  \includegraphics[width=\textwidth]{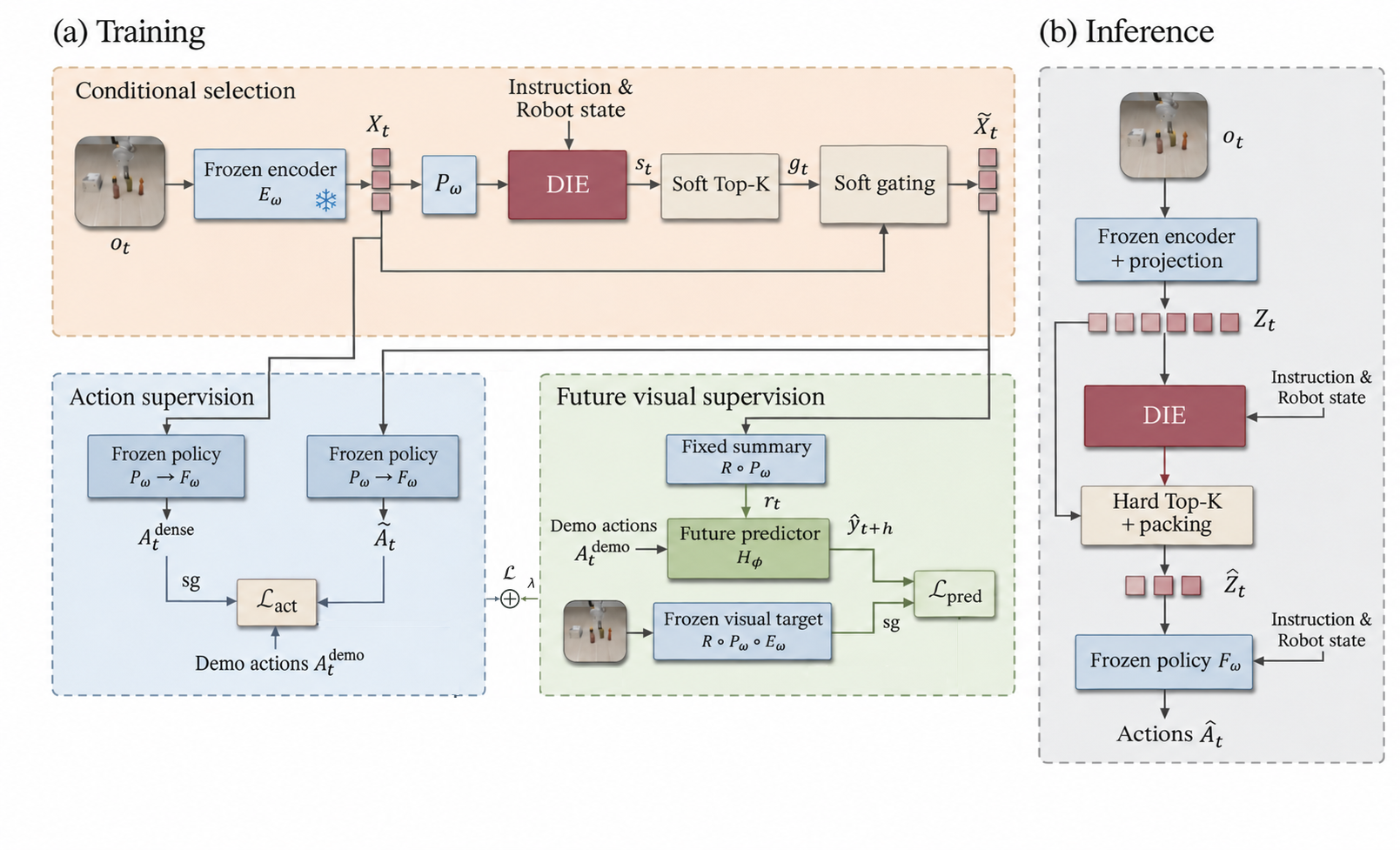}
  \caption{Overview of APS-VLA. A shared visual selector is trained with a weighted combination of action supervision and future visual supervision while the VLA backbone remains frozen. Deployment uses one hard selection and packing step before native action generation; the auxiliary predictor is removed.}
  \label{fig:method}
\end{figure*}

Our contributions are threefold.
\begin{itemize}
\item \textbf{Learning visual selection from demonstration trajectories.} We formulate two complementary objectives for a shared visual selector. Action supervision preserves the information needed to reproduce policy and expert actions, while action-conditioned prediction supervises the same representation with the visual change observed later in the demonstration.
\item \textbf{The Decision Influence Estimator.} We introduce DIE, a lightweight module that scores individual visual tokens in the context of the instruction and robot state (Figs.~\ref{fig:method} and~\ref{fig:die}). Trained with both objectives, DIE converts a fixed budget per camera into a single hard selection step that integrates with the frozen policy's native action generator.
\item \textbf{Extensive simulation and real-world evaluation.} APS-VLA achieves inference speedups of $1.57\times$ on LIBERO, up to $1.52\times$ on RoboCasa365, and $1.47\times$ across five real-world tasks, with task success rates close to the corresponding unpruned baselines.
\end{itemize}

Upon acceptance, we will release the code for reproducing our experiments and all model checkpoints.

\section{Related Work}
\label{sec:related_work}

\noindent\textbf{Visual token compression.} Visual token compression reduces inference cost by shortening the sequence processed by a model. DynamicViT learns to discard tokens progressively, EViT fuses less-attended tokens, and ToMe merges tokens with similar features~\cite{rao2021dynamicvit,liang2022evit,bolya2023tome}. In vision-language models, FastV uses early-layer attention to identify tokens for removal, PruMerge combines selection and merging, and SparseVLM uses text to guide visual sparsification~\cite{chen2024fastv,shang2025prumerge,zhang2025sparsevlm}. CSP reduces the cost of multimodal inference by pruning the key-value cache~\cite{pei2024csp}. For robot control, visual similarity and instruction relevance capture only part of the selection problem. Patches on the same object may be equally relevant to the instruction while providing different information about grasp contact or clearance.

\noindent\textbf{Token pruning and reuse in VLA policies.} Robot policies span discrete action tokens (RT-2 and FAST)~\cite{brohan2023rt2,pertsch2025fast}, action chunks (ACT)~\cite{zhao2023act}, and continuous actions generated by diffusion or flow matching~\cite{chi2023diffusionpolicy,li2024cogact,black2024pi0}. Visual token pruning targets the observation sequence shared by these policies. VLA-Cache reuses computation for visual tokens that remain stable across frames~\cite{xu2025vlacache}. VLA-Pruner combines vision-language prefill attention with an estimate of action-decoding attention from recent steps~\cite{liu2025vlapruner}. VLA-ADP uses end-effector motion to determine when to apply text-guided pruning, while SpecPrune-VLA combines information from preceding steps with current attention and adjusts pruning according to the motion being executed~\cite{pei2026vlaadp,wang2025specprune}. SAFE-Pruner addresses a different source of error: a token assigned low importance in an early layer may be needed deeper in the network. It therefore incorporates attention cues from later layers into pruning decisions~\cite{ma2026safepruner}. These methods use signals available during policy execution. Demonstrations offer an additional signal during training because they record the physical outcome of the action taken from each observation.

\noindent\textbf{Learning visual selection from robot trajectories.}
LightVLA makes token pruning differentiable so that the policy objective can shape the selection rule~\cite{jiang2025lightvla}, while GridS learns spatial coordinates for feature sampling via differentiable interpolation~\cite{feng2026grids}. 
Beyond immediate control signals, prior policy learning studies have shown the value of future visual states: UniPi and SuSIE guide control via predicted videos or image subgoals~\cite{du2023unipi, black2024susie}, GR-1 jointly predicts actions and future images~\cite{wu2023gr1}, and Seer conditions inverse dynamics on predicted visual states~\cite{tian2024seer}. 
Very recently, the concurrent work VLA-ACL~\cite{du2026vlaacl} explores learning a token selector over a frozen VLA backbone using action-consistency supervision from expert demonstrations. This is closely aligned with our insight that visual token importance should be learned through its contribution to action generation. 
Action supervision directly tests whether the retained representation supports the immediate command. 
In long-horizon or unseen compositional tasks, however, distinct interaction states often call for similar motions, where the current action target alone gives limited guidance on which visual details govern how the scene will evolve. 
APS-VLA therefore couples action supervision with action-conditioned visual transition prediction: given the demonstrated actions, the features retained by DIE must support both the current policy output and a later visual representation. 
The prediction head is used only during training, adding no overhead at deployment.

\section{Method}
\label{sec:method}

APS-VLA learns to allocate a limited visual token budget using demonstration trajectories. The selected visual context supports two complementary objectives: producing actions consistent with the policy and demonstrations, and predicting the observed visual transition conditioned on demonstrated actions. Both objectives train a shared selector while the VLA backbone remains frozen. Figure~\ref{fig:method} summarizes the training and deployment procedures.

\subsection{Visual Selection under a Fixed Budget}
\label{sec:selection_interface}

Let $o_t=\{o_t^c\}_{c=1}^{C}$ denote the current multi-view observation, $\ell$ the instruction, and $p_t$ the normalized proprioception. A frozen visual encoder $E_\omega$ and projector $P_\omega$ produce
\begin{equation}
  X_t^c=E_\omega(o_t^c), \qquad Z_t^c=P_\omega(X_t^c).
  \label{eq:visual_features}
\end{equation}
The projector is the identity when selection operates directly on the backbone's visual-prefix representation. We denote the action-generating policy by $F_\omega$.

A selector $S_\theta$ scores visual tokens using only current inputs:
\begin{equation}
  s_t=S_\theta(Z_t,\ell,p_t).
  \label{eq:selection_interface}
\end{equation}
For a total budget $K$, fixed per-view budgets $k_c$ satisfy $\sum_c k_c=K$. During training, the scores determine differentiable gates $g_{t,i}^c\in[0,1]$, with $\sum_i g_{t,i}^c=k_c$. These gates construct a shared visual representation $\widetilde X_t$, as detailed in Sec.~\ref{sec:die}. Training preserves the full token layout for differentiation; deployment uses the scores to select and physically pack a smaller token set.

We optimize the selector and a training-time predictor $H_\phi$ with
\begin{equation}
  \mathcal L=\mathcal L_{\mathrm{act}}+\lambda\mathcal L_{\mathrm{pred}},
  \label{eq:total_loss}
\end{equation}
where $\lambda$ weights the prediction objective. Language and proprioception condition selection but are not pruned. All backbone parameters $\omega$ remain fixed.

\begin{figure}[t]
  \centering
  \includegraphics[width=\columnwidth]{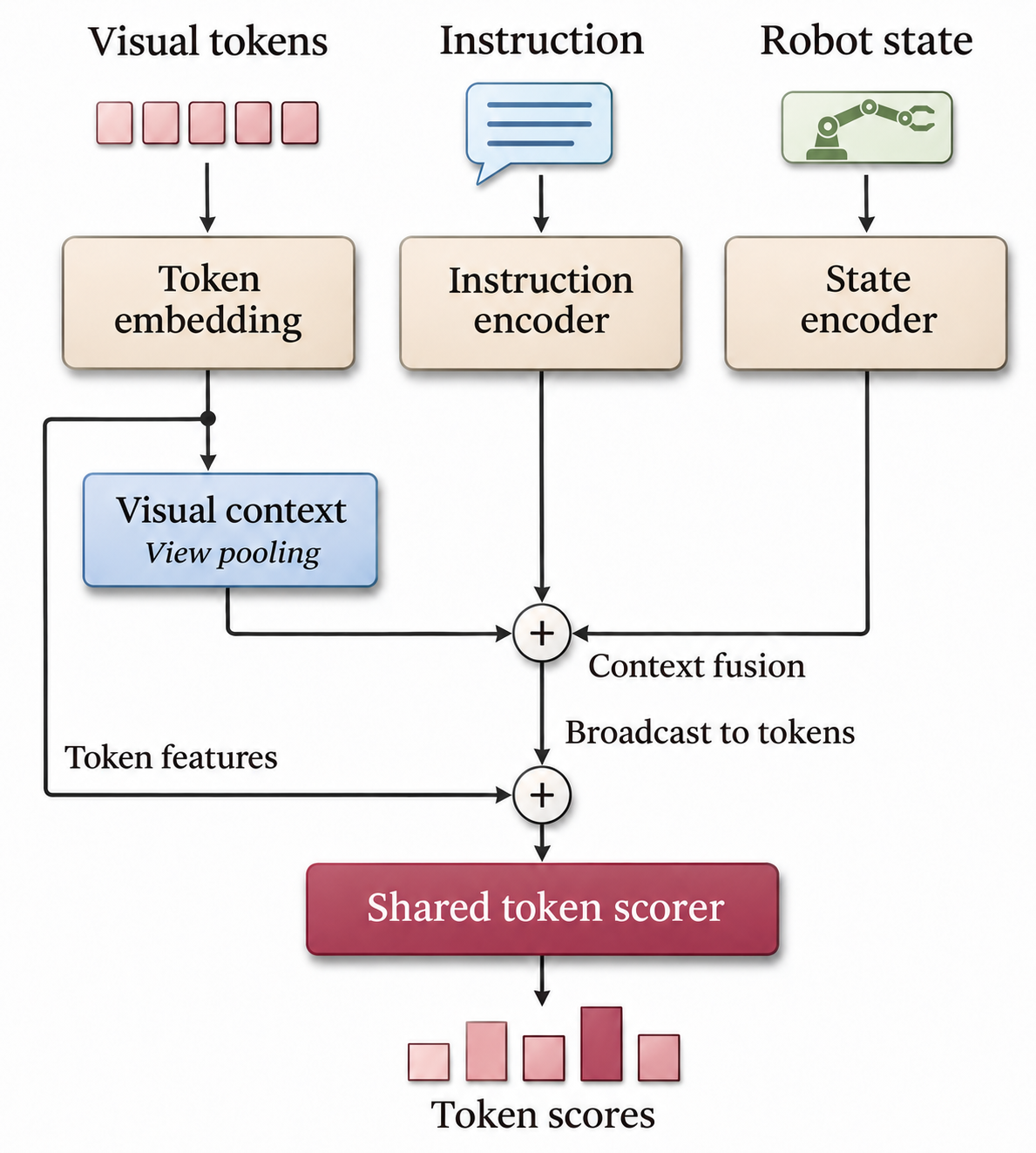}
  \caption{Decision Influence Estimator (DIE). Visual context, the instruction, and robot state are fused and added to each visual token feature. A shared scorer then produces one score per token. Robot state includes proprioceptive inputs such as end-effector pose and gripper state.}
  \label{fig:die}
\end{figure}

\subsection{Joint Supervision from Demonstration Trajectories}
\label{sec:joint_supervision}
\label{sec:action_supervision}
\label{sec:future_prediction}

A demonstration provides both recorded actions and the observations reached through those actions. We use these signals to supervise the control output and predictive content of the selected visual context.

\paragraph{Action supervision.}
The action branch evaluates the gated context through the frozen policy. It uses two targets: the policy's full-context prediction and the demonstrated action chunk. Let $\mathcal T_h$ extract the first $h$ actions in the task's physical action dimensions from the backbone's native output. We compute
\begin{equation}
  \begin{aligned}
    A_t^{\mathrm{dense}} &= \operatorname{sg}\!\left[\mathcal T_h F_\omega(P_\omega(X_t),\ell,p_t)\right],\\
    \widetilde A_t &= \mathcal T_h F_\omega(P_\omega(\widetilde X_t),\ell,p_t),
  \end{aligned}
  \label{eq:action_predictions}
\end{equation}
where $\operatorname{sg}(\cdot)$ denotes stop-gradient. For stochastic action generation, the dense and gated branches share the same sampling noise.

Given the normalized demonstration chunk $A_t^{\mathrm{demo}}\in\mathbb R^{h\times d_a}$, we minimize
\begin{equation}
  \mathcal L_{\mathrm{act}}=
  \frac{\|\widetilde A_t-A_t^{\mathrm{dense}}\|_1+\|\widetilde A_t-A_t^{\mathrm{demo}}\|_1}{2h d_a}.
  \label{eq:action_loss}
\end{equation}
The dense target encourages preservation of the pretrained policy's behavior, while the demonstration target anchors the output to recorded actions. Gradients pass through the frozen policy to the selector without updating the backbone.

\paragraph{Visual transition supervision.}
An action-conditioned predictor uses the gated current representation and demonstrated actions to predict a summary of the corresponding future observation.

We define a fixed summary operator $R$. Within each view, it averages projected tokens over a $2\times2$ spatial grid, applies a fixed Gaussian projection to 64 dimensions per cell, and performs non-affine layer normalization. The resulting features are concatenated in a fixed camera and spatial order. The prediction and target are
\begin{align}
  r_t &= R(P_\omega(\widetilde X_t)),\label{eq:current_summary}\\
  \widehat y_{t+h} &= H_\phi\!\left(r_t,\operatorname{vec}(A_t^{\mathrm{demo}})\right),\label{eq:future_prediction}\\
  y_{t+h} &= \operatorname{sg}\!\left[R(P_\omega(E_\omega(o_{t+h})))\right].\label{eq:future_target}
\end{align}
Here, $H_\phi$ is a two-layer predictor, and $o_{t+h}$ is the observation reached after the corresponding demonstrated actions. We minimize
\begin{equation}
  \mathcal L_{\mathrm{pred}}=\frac{1}{d_y}\|\widehat y_{t+h}-y_{t+h}\|_2^2,
  \label{eq:prediction_loss}
\end{equation}
where $d_y$ is the target dimension.

The demonstrated actions specify the transition being predicted. Visual information reaches the predictor only through the gated current summary, allowing the prediction loss to supervise which current features the selector retains.

Both losses update the shared selector; the prediction loss also updates $H_\phi$. Demonstrated actions enter the auxiliary predictor only during training, and future observations supply detached targets. Neither is an input to the deployed selector.

\subsection{Implementing Differentiable Visual Selection}
\label{sec:die}

\paragraph{Task-conditioned scoring.}
The Decision Influence Estimator (DIE), illustrated in Fig.~\ref{fig:die}, combines local token features with shared visual, language, and proprioceptive context. For each visual token, we apply layer normalization and a learned linear projection, then add camera and patch embeddings:
\begin{equation}
  u_{t,i}^{c}=W_z\operatorname{LN}(Z_{t,i}^{c})+b_z+e_{\mathrm{cam}}^c+e_{\mathrm{patch}}^i.
  \label{eq:token_embedding}
\end{equation}

Mean and maximum features summarize each view. A learned affine mapping produces the visual condition:
\begin{equation}
  c_t^{\mathrm{vis}}=\phi_v\!\left(\operatorname*{Concat}_{c=1}^{C}\!\left[\operatorname*{Mean}_i u_{t,i}^{c},\operatorname*{Max}_i u_{t,i}^{c}\right]\right).
  \label{eq:visual_context}
\end{equation}
Instruction embeddings come from the frozen backbone's token embedding table. A learned projection, positional embeddings, and a token-wise nonlinear encoder are followed by averaging over valid instruction tokens to obtain $c_t^\ell$. A separate MLP maps normalized proprioception to $c_t^p$.

The conditions are added and broadcast to every token:
\begin{equation}
  \begin{aligned}
    q_t &= c_t^{\mathrm{vis}}+c_t^\ell+c_t^p,\\
    s_{t,i}^{c} &= f_s(u_{t,i}^{c}+q_t).
  \end{aligned}
  \label{eq:conditional_scores}
\end{equation}
The shared scoring network $f_s$ consists of layer normalization and a two-layer MLP with a SiLU activation. Visual features, instruction embeddings, and proprioception are detached at the selector input.

\paragraph{Budget-constrained gates.}
Differentiable token pruning has been explored in prior work~\cite{rao2021dynamicvit,jiang2025lightvla}; here we use the following budget-constrained relaxation.
To train the selector through the objectives in Sec.~\ref{sec:joint_supervision}, we convert scores into differentiable gates:
\begin{equation}
  g_{t,i}^{c}=F_{\mathrm{Lap}}\!\left(\frac{s_{t,i}^{c}-b_t^c}{\alpha}\right),
  \qquad \sum_i g_{t,i}^{c}=k_c,
  \label{eq:soft_selection}
\end{equation}
where $F_{\mathrm{Lap}}$ is the standard Laplace cumulative distribution function and $\alpha>0$ controls selection sharpness. The threshold $b_t^c$ is solved separately for each view to satisfy its budget. Gradients are obtained by implicit differentiation of the budget constraint, and the temperature is annealed during training.

Continuous token gating has been explored in DiffPrune~\cite{he2026diffprune} and the contemporaneous VLA-ACL~\cite{du2026vlaacl}. We construct the gated representation using black-reference features $B_i^c$ and detach the reference coefficient during backpropagation:
\begin{equation}
  \widetilde X_{t,i}^{c}=g_{t,i}^{c}X_{t,i}^{c}
  +\sqrt{1-\operatorname{sg}(g_{t,i}^{c})^2}\,B_i^c.
  \label{eq:black_reference_gating}
\end{equation}
Both $X_t^c$ and $B^c$ are detached. The frozen projector and policy remain differentiable with respect to their inputs.

This representation supplies both supervision branches. Consequently, action matching and visual transition prediction shape the same token scores through the same gates.

\subsection{Deployment with Compact Visual Inputs}
\label{sec:hard_inference}

At deployment, the selector scores each newly observed input once and applies hard selection within each view:
\begin{equation}
  \mathcal I_t^c=\operatorname{TopK}_{k_c}(s_t^c).
  \label{eq:hard_selection}
\end{equation}
Ties are resolved deterministically by original token index. Retained tokens are restored to their original camera and patch order and physically packed into a compact sequence:
\begin{equation}
  \widehat Z_t=\operatorname{Pack}\!\left(\{Z_{t,i}^{c}:i\in\mathcal I_t^c\}_{c=1}^{C}\right).
  \label{eq:token_packing}
\end{equation}
Packing preserves camera and spatial identities. Position identifiers, attention masks, and action-output indexing follow the backbone's consistent packing rules. Language and proprioception remain intact.

The frozen policy then generates actions using its native action-generation procedure:
\begin{equation}
  \widehat A_t=F_\omega(\widehat Z_t,\ell,p_t).
  \label{eq:compact_inference}
\end{equation}
As shown in Fig.~\ref{fig:method}(b), deployment requires only the selector and the frozen policy. The dense teacher branch, black-reference gating, and future predictor are removed.

\section{Experiments}
\label{sec:experiments}

We study APS-VLA in simulation and real-world manipulation. Simulation experiments consider OpenVLA-OFT~\cite{kim2025oft} on LIBERO~\cite{liu2023libero} and both $\pi_{0.5}$~\cite{intelligence2025pi05} and XR-1~\cite{xiaomi2026xr1} on RoboCasa365~\cite{nasiriany2024robocasa,nasiriany2026robocasa365}, covering parallel action prediction and flow-matching policies.

\subsection{Experimental Settings}
\label{sec:experimental_settings}

We freeze the VLA backbone and train the token selector and conditioning module with an auxiliary future prediction head. Unless otherwise stated, training uses AdamW~\cite{loshchilov2019adamw} with a learning rate of $10^{-4}$, a batch size of eight, and 5,000 updates. The default future prediction horizon is $h=8$, with an auxiliary loss weight of $0.1$; bounded training prefixes and weight variations are specified in Sec.~\ref{sec:ablation_study}. The future prediction head is removed at deployment.

\paragraph{Evaluation metrics.}
We report task success rate, model inference latency, and inference speedup. Latency includes visual encoding, token selection and packing, language-backbone computation, and action prediction.

\subsection{Simulation Experiments}
\label{sec:simulation_experiments}

\subsubsection{OpenVLA-OFT on LIBERO}
\label{sec:libero_experiments}

LIBERO~\cite{liu2023libero} comprises four task suites: Spatial, Object, Goal, and Long. OpenVLA-OFT~\cite{kim2025oft} receives 512 visual tokens from two camera views. We evaluate pruning ratios of 50\%, 75\%, and 87.5\%, corresponding to 256, 128, and 64 retained tokens, respectively.

\begin{table*}[t]
  \centering
  \caption{Comparison with the baseline and representative VLA methods on LIBERO. Success rates are reported in percentages. All methods, including APS-VLA and the unpruned OpenVLA-OFT baseline, are reproduced on NVIDIA A800 GPUs under the same evaluation conditions.}
  \label{tab:libero_comparison}
  % Provenance update, 2026-11-10: user confirms all compared methods were reproduced on A800 under the same conditions.
  % User reports the additional run records were not saved; this edit does not independently verify those runs or alter table values.
  \small
  \setlength{\tabcolsep}{4pt}
  \renewcommand{\arraystretch}{1.12}
  \resizebox{\textwidth}{!}{%
  \begin{tabular}{lccc*{7}{r}}
    \hline
    Method & Decoding & CKPT & Pruning ratio & Spatial $\uparrow$ & Object $\uparrow$ & Goal $\uparrow$ & Long $\uparrow$ & Average $\uparrow$ & Model latency (ms) $\downarrow$ & Speedup $\uparrow$ \\
    \hline
    OpenVLA~\cite{kim2024openvla} & AR & OpenVLA (7B) & --- & 84.70 & 88.40 & 79.20 & 53.70 & 76.50 & --- & --- \\
    OpenVLA-OFT~\cite{kim2025oft} & PD & OFT (7B) & 0\% & 98.00 & \textbf{99.00} & 97.00 & 94.00 & \textbf{97.00} & 107.95 & $1.00\times$ \\
    CSP (+OFT)~\cite{pei2024csp} & PD & OFT (7B) & --- & 84.70 & 82.20 & 77.10 & 74.30 & 79.60 & --- & --- \\
    FastV (+OFT)~\cite{chen2024fastv} & PD & OFT (7B) & --- & 96.80 & 81.00 & 96.40 & 73.00 & 86.80 & 93.68 & $1.15\times$ \\
    VLA-Cache~\cite{xu2025vlacache} & PD & OFT (7B) & --- & 97.30 & 97.50 & 96.30 & 94.40 & 96.38 & 92.60 & $1.17\times$ \\
    VLA-ADP~\cite{pei2026vlaadp} & PD & OFT (7B) & --- & 97.40 & 98.00 & 96.40 & 91.20 & 95.75 & 94.17 & $1.15\times$ \\
    VLA-ACL~\cite{du2026vlaacl} & PD & OFT (7B) & --- & 96.40 & 98.00 & 96.60 & 92.30 & 95.83 & 71.49 & $1.51\times$ \\
    SAFE-Pruner~\cite{ma2026safepruner} & PD & OFT (7B) & --- & 98.00 & 98.00 & 96.20 & 93.40 & 96.40 & 73.30 & $1.47\times$ \\
    SpecPrune-VLA~\cite{wang2025specprune} & PD & OFT (7B) & --- & 97.40 & 95.80 & \textbf{97.70} & 93.40 & 96.10 & 73.94 & $1.46\times$ \\
    VLA-Pruner~\cite{liu2025vlapruner} & PD & OFT (7B) & --- & 93.50 & 96.20 & 95.20 & 90.20 & 93.80 & 74.60 & $1.45\times$ \\
    % LightVLA row: user-reported measurements, 2026-11-04; 78 retained of 512 tokens.
    % Mean success = 95.45; latency = 107.94801125833334 / 1.46 = 73.93699401255708 ms.
    LightVLA~\cite{jiang2025lightvla}\hphantom{ (78 tokens)} & PD & OFT (7B) & --- & 97.60 & 94.20 & 95.40 & 94.60 & 95.45 & 73.94 & $1.46\times$ \\
    \hline
    APS-VLA & PD & OFT (7B) & 50\% & \textbf{98.33} & 97.67 & 95.33 & \textbf{95.00} & \underline{96.58} & 81.39 & $1.33\times$ \\
    APS-VLA & PD & OFT (7B) & 75\% & 96.33 & 97.67 & 96.00 & 93.00 & 95.75 & 68.75 & $1.57\times$ \\
    APS-VLA & PD & OFT (7B) & 87.5\% & 96.33 & 96.33 & 92.67 & 90.33 & 93.92 & \textbf{66.21} & $\mathbf{1.63\times}$ \\
    \hline
  \end{tabular}%
  }
  \vspace{3pt}
  \parbox{\textwidth}{\footnotesize AR: autoregressive decoding. PD: parallel decoding. APS-VLA model inference speedups are measured using sampled calls and averaged across the four LIBERO suites.}
\par\medskip
  \centering
  \includegraphics[width=.70\textwidth]{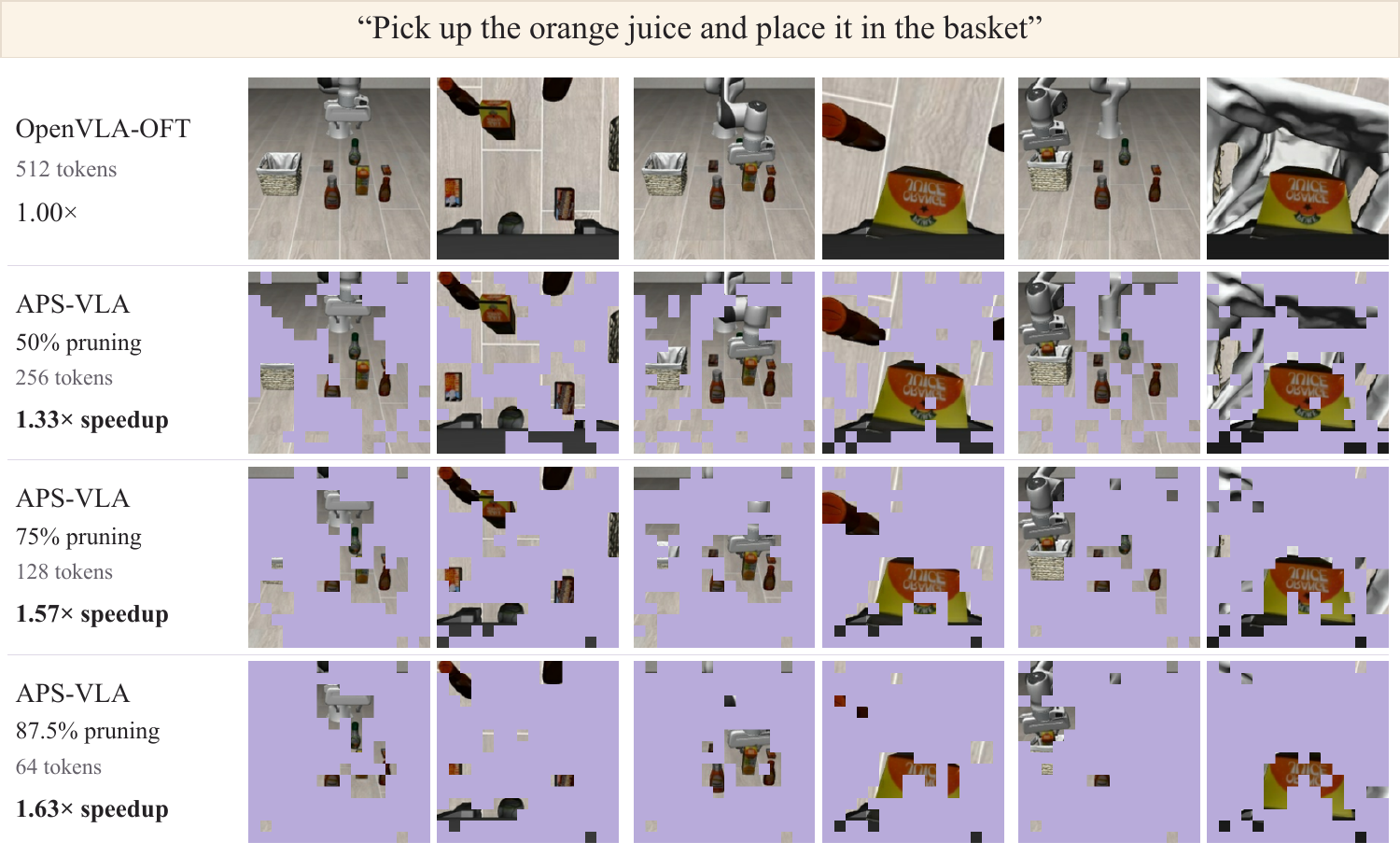}
  \captionof{figure}{Visual token selection at three pruning ratios on the same LIBERO-Object observations. Purple denotes discarded patches; speedups refer to policy inference in Table~\ref{tab:libero_comparison}.}
  \label{fig:pruning_visualization}
\end{table*}

At 75\% pruning, policy inference latency drops significantly to 68.75~ms (a $1.57\times$ speedup), while the mean LIBERO success rate remains at 95.75\%---only 1.25 percentage points below the unpruned baseline (Table~\ref{tab:libero_comparison}). This establishes a favorable operating point where redundant background patches are successfully discarded without sacrificing task-critical details.

However, increasing the pruning ratio to 87.5\% (retaining only 64 tokens) yields diminishing returns in latency (66.21~ms) while incurring a sharper drop in success rate (93.92\%). This non-linear performance degradation indicates that dropping more than 75\% of the visual tokens begins to strip away fine-grained spatial cues essential for precise contact estimation and multi-stage coordination.

Figure~\ref{fig:pruning_visualization} compares the hard token selections on identical observations at the three pruning ratios.

\subsubsection{\texorpdfstring{$\pi_{0.5}$}{pi0.5} and XR-1 on RoboCasa365}
\label{sec:robocasa_experiments}

To test whether APS-VLA preserves the skills needed for longer task sequences across VLA architectures, we evaluate $\pi_{0.5}$~\cite{intelligence2025pi05} and Xiaomi-Robotics-1 (XR-1)~\cite{xiaomi2026xr1} on RoboCasa365~\cite{nasiriany2026robocasa365}. We compare against FastV, SAFE-Pruner, and VLA-ACL, three pruning methods previously evaluated with the flow-matching policy $\pi_{0.5}$~\cite{ma2026safepruner,du2026vlaacl}. We use the reported unpruned $\pi_{0.5}$ results and measure the unpruned XR-1 baseline in our setup.

\begin{table*}[t]
  \centering
  \caption{Visual token pruning on RoboCasa365. We report success rates (\%) on seen atomic (A-S), seen composite (C-S), and unseen composite (C-U) tasks, policy inference latency (ms), and speedup relative to each unpruned policy. APS-VLA averages success rates and speedups over the 256- and 128-token budgets. All methods are reproduced on NVIDIA A800 GPUs under the same evaluation conditions.}
  \label{tab:robocasa_results}
  \small
  \setlength{\tabcolsep}{3.5pt}
  \renewcommand{\arraystretch}{1.08}
  \begin{tabular*}{\textwidth}{@{\extracolsep{\fill}}lrrrrr@{\hspace{1em}}rrrrr@{}}
    \hline
    & \multicolumn{5}{c}{$\pi_{0.5}$} & \multicolumn{5}{c}{XR-1} \\
    \cline{2-6}\cline{7-11}
    Method & A-S & C-S & C-U & Latency & Speedup & A-S & C-S & C-U & Latency & Speedup \\
    \hline
    Unpruned & \textbf{39.60} & \textbf{7.10} & \underline{1.20} & 70.29--70.30 & $1.00\times$ & \textbf{80.23} & \textbf{57.10} & \textbf{32.10} & 217.25 & $1.00\times$ \\
    FastV~\cite{chen2024fastv} & 36.47 & 5.67 & 0.30 & 56.24 & $1.25\times$ & 78.41 & 52.67 & 25.67 & 159.74 & $1.36\times$ \\
    SAFE-Pruner~\cite{ma2026safepruner} & 37.50 & 6.33 & 0.87 & 55.35 & $1.27\times$ & 78.10 & 49.23 & 25.92 & 164.58 & $1.32\times$ \\
    VLA-ACL~\cite{du2026vlaacl} & 38.97 & 6.33 & 0.77 & 52.46 & $1.34\times$ & 79.93 & 52.10 & 29.33 & 149.83 & $1.45\times$ \\
    \hline
    APS-VLA & \underline{39.15} & \underline{6.71} & \textbf{1.21} & \textbf{51.50} & $\mathbf{1.37\times}$ & 79.77 & \underline{55.67} & \underline{30.97} & \textbf{142.93} & $\mathbf{1.52\times}$ \\
    \hline
  \end{tabular*}
\end{table*}

APS-VLA achieves higher success rates than the three other pruning methods on both composite splits (Table~\ref{tab:robocasa_results}). On XR-1, its gains over the strongest pruning baseline are 3.00 and 1.64 percentage points on seen and unseen compositions, respectively, at a $1.52\times$ mean speedup. On seen atomic tasks, its mean success is within half a percentage point of the unpruned policy on both backbones.

\raggedbottom
\subsection{Ablation Study}
\label{sec:ablation_study}

We conduct ablation experiments using OpenVLA-OFT on LIBERO, with three random seeds for each experiment.

\paragraph{Future visual supervision.}
We compare action supervision alone (A) with additional future visual supervision (B), keeping all other training and deployment settings unchanged.

Future visual supervision improves success rates by 1.41, 1.17, and 1.17 percentage points at token budgets of 256, 128, and 64, respectively (Table~\ref{tab:future_ablation}). These consistent gains indicate that future visual supervision helps the selector retain information that is more useful for task execution.

% Provenance: A values are updated from user-reported latest measurements on 2026-11-03.
% B values retain the existing development aggregates from
% results/formal_main_v2/speedup_sample_20261021/three_budget_speed_accuracy_summary.json.
% That summary still flags full terminal acceptance as pending for Goal
% k256/seed23 and k64/seed23,41. These are not sealed final-test results.
\begin{table}[H]
  \centering
  \caption{Mean development success (\%) across four LIBERO suites at different visual token budgets.}
  \label{tab:future_ablation}
  \small
  \setlength{\tabcolsep}{4pt}
  \renewcommand{\arraystretch}{1.10}
  \begin{tabular}{ccrr}
    \hline
    Tokens & Pruning & A & B \\
    \hline
    256 & 50\% & 95.17 & 96.58 \\
    128 & 75\% & 94.58 & 95.75 \\
    64 & 87.5\% & 92.75 & 93.92 \\
    \hline
  \end{tabular}
\end{table}

% Declare wide real-world floats on the preceding page.
\begin{figure*}[!t]
  \centering
  \includegraphics[width=\textwidth,height=260pt,keepaspectratio]{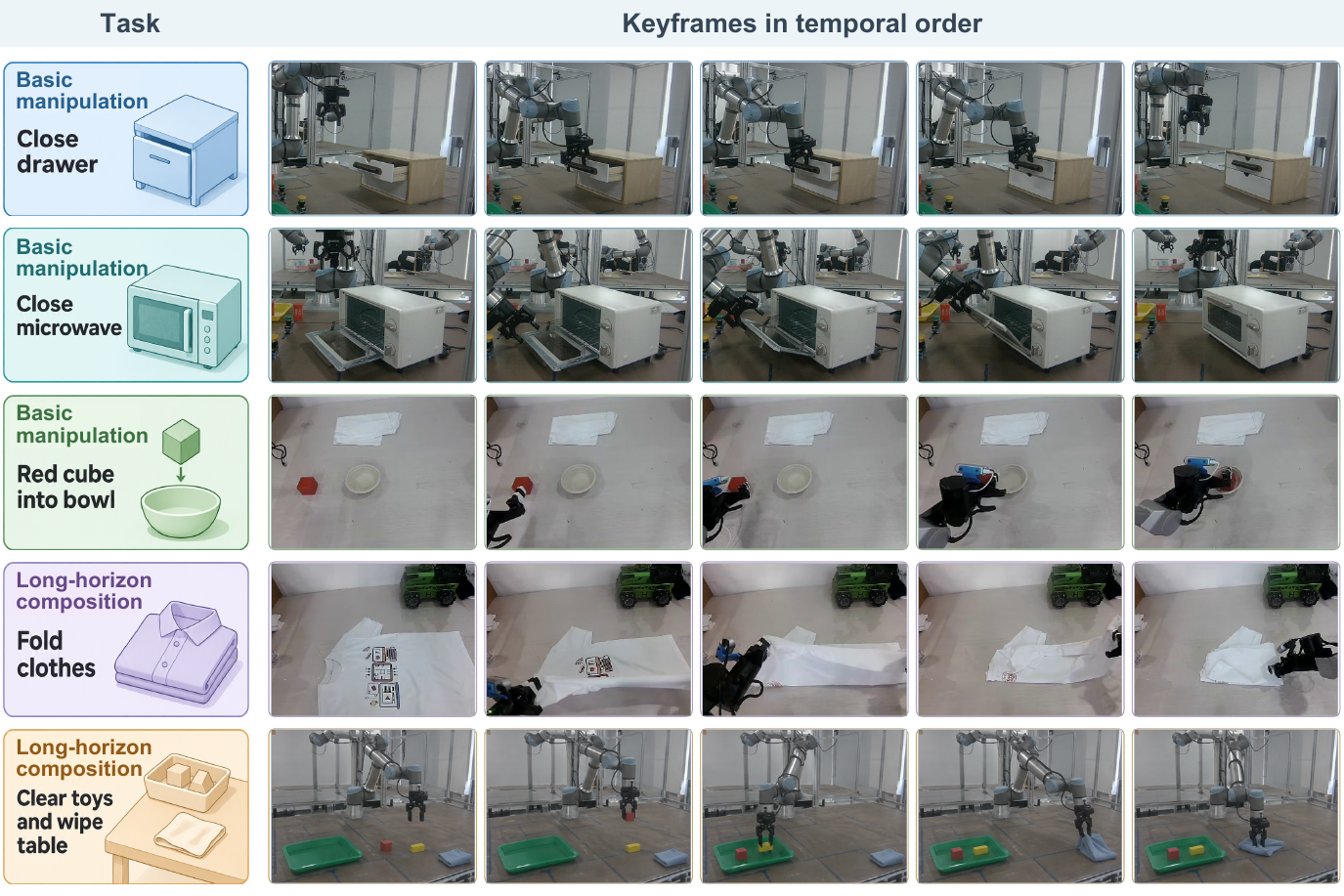}
  \caption{Representative keyframes from five real-world manipulation tasks. Frames within each task are ordered chronologically, highlighting object interactions and transitions between task stages.}
  \label{fig:real_world_tasks}
\end{figure*}

% Queue the wide real-world table early; its reading order follows the two remaining ablation tables.
\addtocounter{table}{2}
\begin{table*}[!t]
  \centering
  \caption{Real-world task success rates (\%) and policy inference latency (ms). Each method is evaluated over 40 trials per task. Speedup is relative to the corresponding unpruned $\pi_{0.5}$ baseline.}
  \label{tab:real_world_results}
  \small
  \setlength{\tabcolsep}{8pt}
  \renewcommand{\arraystretch}{1.10}
  \begin{tabular}{lccccc}
    \hline
    & \multicolumn{2}{c}{Success rate (\%) $\uparrow$} & \multicolumn{2}{c}{Inference latency (ms) $\downarrow$} & \\
    Task & $\pi_{0.5}$ & APS-VLA & $\pi_{0.5}$ & APS-VLA & Speedup $\uparrow$ \\
    \hline
    Close Drawer & 80.00 & \textbf{82.50} & 131.25 & \textbf{91.25} & $1.44\times$ \\
    Close Microwave & \textbf{75.00} & 72.50 & 134.50 & \textbf{92.50} & $1.45\times$ \\
    Red Cube into Bowl & \textbf{72.50} & \textbf{72.50} & 132.75 & \textbf{90.75} & $1.46\times$ \\
    Fold Clothes & \textbf{37.50} & \textbf{37.50} & 142.50 & \textbf{95.25} & $1.50\times$ \\
    Clear Toys and Wipe Table & 35.00 & \textbf{40.00} & 145.25 & \textbf{96.50} & $1.51\times$ \\
    \hline
    Average & 60.00 & \textbf{61.00} & 137.25 & \textbf{93.25} & $1.47\times$ \\
    \hline
  \end{tabular}
  \vspace{3pt}
  \parbox{\textwidth}{\footnotesize Averages are task-weighted equally. Aggregate speedup is the ratio of average inference latencies.}
\end{table*}
\addtocounter{table}{-3}

\paragraph{Temporal correspondence of future targets.}
On LIBERO-Long, we fix $K=128$, $h=8$, and $\lambda=0.1$ and train matched selectors with aligned or shuffled future targets for 2,000 updates. The shuffled control permutes future targets across different episodes of the same task, while keeping all other training settings unchanged. Table~\ref{tab:temporal_correspondence} reports task-macro teacher-action $L_1$ and closed-loop rollout success across three seeds. Aligned targets yield lower action error and higher success at every stage, reducing $L_1$ by 37.8\% and raising rollout success by 5.00 percentage points (94.33\% vs.\ 89.33\%) at 2,000 updates.

\begin{table}[H]
  \centering
  \caption{Temporal-target control on LIBERO-Long ($K=128$): teacher-action $L_1$ (task-weighted, averaged over three seeds) and closed-loop success rate.}
  \label{tab:temporal_correspondence}
  \footnotesize
  \setlength{\tabcolsep}{3pt}
  \renewcommand{\arraystretch}{1.10}
  \resizebox{\columnwidth}{!}{%
  \begin{tabular}{crrrr}
    \hline
    Updates & \shortstack{Aligned\\$L_1\downarrow$} & \shortstack{Shuffled\\$L_1\downarrow$} & \shortstack{Aligned\\Succ. (\%) $\uparrow$} & \shortstack{Shuffled\\Succ. (\%) $\uparrow$} \\
    \hline
    500 & 0.016451 & 0.022377 & 91.33 & 87.00 \\
    1,000 & 0.015625 & 0.021379 & 93.67 & 88.33 \\
    1,500 & 0.013377 & 0.020357 & 94.00 & 89.00 \\
    2,000 & \textbf{0.011973} & 0.019251 & \textbf{94.33} & 89.33 \\
    \hline
  \end{tabular}%
  }
\end{table}

\paragraph{Auxiliary-weight sensitivity.}
We evaluate $\lambda\in\{0.03,0.1,0.3\}$ on LIBERO-Long with $K=128$, $h=8$, and 1,000 training updates. The $\lambda=0.1$ runs reuse checkpoints from the first 1,000 updates of the matched aligned-target controls; the other six selectors are trained from initialization. For each run, we select among updates 0, 500, and 1,000 by minimum task-macro hard-selection teacher-action L1, breaking ties in favor of the earlier checkpoint. Each of the nine selected models is evaluated on the same 100 development initial states, yielding 900 completed rollouts. Table~\ref{tab:lambda_sensitivity} gives each seed's success rate and the mean with sample standard deviation across the three training seeds.

Mean success ranges from 92.00\% to 93.67\% across the three weights, a maximum difference of 1.67 percentage points, indicating stable performance within the tested range. The highest mean success is achieved with $\lambda=0.1$, supporting its use as the default setting.

\begin{table}[H]
  \centering
  \caption{Auxiliary-weight sensitivity on LIBERO-Long ($K=128$, $h=8$). Development success (\%) on 100 shared initial states per model; the final column is mean $\pm$ sample standard deviation over training seeds, in percentage points.}
  \label{tab:lambda_sensitivity}
  \small
  \setlength{\tabcolsep}{4pt}
  \renewcommand{\arraystretch}{1.10}
  \begin{tabular}{crrrc}
    \hline
    $\lambda$ & Seed 17 & Seed 23 & Seed 41 & Mean $\pm$ SD \\
    \hline
    0.03 & 91 & 94 & 91 & $92.00\pm1.73$ \\
    0.1 & 94 & 92 & 95 & $\mathbf{93.67\pm1.53}$ \\
    0.3 & 92 & 91 & 94 & $92.33\pm1.53$ \\
    \hline
  \end{tabular}
\end{table}

\subsection{Real-World Experiments}
\label{sec:real_world_experiments}

We evaluate five real-world tasks on YHRG THESEUS-S1 and dual-UR5 platforms, each equipped with three cameras (Fig.~\ref{fig:real_world_tasks}). APS-VLA and the unpruned baseline use the same task-fine-tuned $\pi_{0.5}$ checkpoints, trained from several hundred demonstrations per task. Each method is evaluated over 40 trials per task, totaling 400 robot executions, on a workstation with two NVIDIA RTX 4090 GPUs.

As shown in Table~\ref{tab:real_world_results}, APS-VLA consistently reduces inference latency across all five tasks, achieving an aggregate speedup of $1.47\times$ while retaining comparable observed success rates. This demonstrates a favorable trade-off between efficiency and task performance across both basic and multi-stage real-world manipulation.

\section{Conclusion}
We presented APS-VLA, which learns visual token selection through action and visual-transition supervision while keeping the VLA backbone frozen. Deployment uses only current inputs and preserves native action generation. Experiments on LIBERO, RoboCasa365, and two real-world platforms demonstrate reduced inference latency with largely preserved task success. Retaining 128 tokens yields a $1.57\times$ speedup on LIBERO, while mean real-world latency falls from 137.25 to 93.25~ms. Future work will explore adaptive token allocation across cameras and temporal reuse of visual cues under a fixed inference budget.

% Reproduction details belong in the appendix; the qualitative token-selection
% visualization uses recorded observations and previously measured sampled O2A.
% Current abstract and LIBERO numbers are development results, not sealed
% final-test results. Sampled O2A measurements retain all timing spikes.

\section*{LLM Usage Statement}
We used OpenAI ChatGPT and Codex (GPT-6 Astra and GPT-6 Sol) to assist with literature review and discussions of experimental analysis. This assistance also included reimplementing comparison methods from their published descriptions when official code or pretrained checkpoints were unavailable. We used OpenAI's image generation tool to create the decorative task icons in Figure~\ref{fig:real_world_tasks}.

{\small
\bibliographystyle{ieeenat_fullname}
\bibliography{main}
}
\end{document}